\documentclass[letterpaper, 10 pt, conference]{ieeeconf}
\IEEEoverridecommandlockouts
\let\labelindent\relax
\usepackage{cite}
\usepackage{amsmath,amssymb,amsfonts}
\usepackage[ruled,linesnumbered,boxed,noend]{algorithm2e}
\usepackage{graphicx}
\usepackage{textcomp}
\usepackage{xcolor}
\usepackage{booktabs}
\usepackage{tabularx} 
\usepackage{multirow} 
\usepackage[colorlinks, linkcolor=black, anchorcolor=red, citecolor=black]{hyperref}
\usepackage{threeparttable}
\usepackage{enumitem}
\usepackage[font={small}]{caption}
\usepackage{balance}
\usepackage{textcomp}
\usepackage{stfloats}
\usepackage{siunitx}
\usepackage{bm}
\usepackage{epstopdf}

\title{\LARGE \bf 
OVAL: Open-Vocabulary Augmented Memory Model for Lifelong Object Goal Navigation
}
\author{ Jiahua Pei$^{*}$, Yi Liu$^{*}$, Guoping Pan, Yuanhao Jiang, Houde Liu$^{\dag}$, Xueqian Wang
}

\begin{document}

\twocolumn[{%
\renewcommand\twocolumn[1][]{#1}%
\maketitle
\begin{center}
  \includegraphics[width=\linewidth]{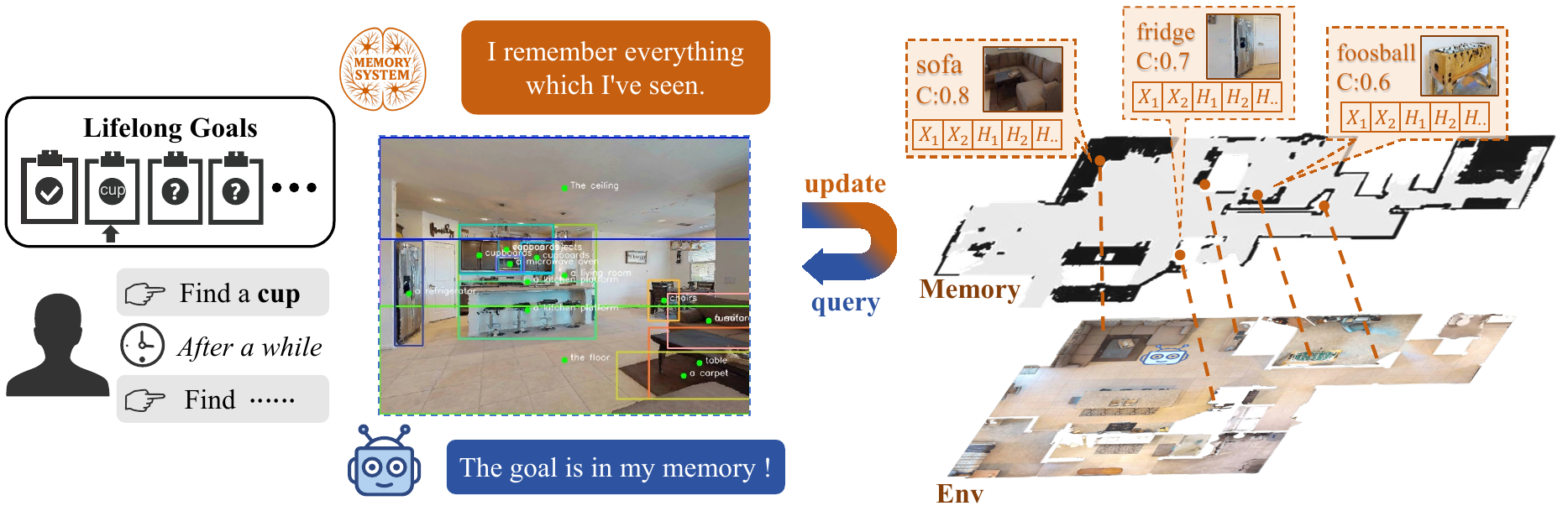}
  \captionof{figure}{\textbf{Motivation:} In hotel scenarios, navigation requires sequentially achieving lifelong open-vocabulary goals in different unseen room scenes. To achieve Lifelong ObjectNav, robots must demonstrate the capability for structured memorization of all scene information in unknown environments, enabling rapid query of target objects when confronted with continual open-vocabulary ObjectNav tasks.}
  \label{first}
\end{center}
\vspace{0.4cm}
}]

\makeatletter{\renewcommand*{\@makefnmark}{}
\footnotetext{* contributed equally; $^\dag$ corresponding authors.\par This work was supported by the Shenzhen Science and Technology Program (Grant No. RCJC20210706091946001) and the Shenzhen Science and Technology Program (Grant No. ZDCY20250901104207008).\par Jiahua Pei, Yi Liu, Guoping Pan, Yuanhao Jiang, Houde Liu, Xueqian Wang are with Shenzhen International Graduate School, Tsinghua University, Shenzhen 518055, China, E-mail: \tt\{pjh24@mails., yiliu24@mails., pgp23@mails., jiang-yh24@mails., liu.hd@sz., wang.xq@sz.\}tsinghua.edu.cn}

\begin{abstract}

Object Goal Navigation (ObjectNav) refers to an agent navigating to an object in an unseen environment, which is an ability often required in the accomplishment of complex tasks. While existing methods demonstrate proficiency in isolated single object navigation, their limitations emerge in the restricted applicability of lifelong memory representations, which ultimately hinders effective navigation toward continual targets over extended periods. To address this problem, we propose OVAL, a novel lifelong open-vocabulary memory framework, which enables efficient and precise execution of long-term navigation in semantically open tasks. Within this framework, we introduce memory descriptors to facilitate structured management of the memory model. Additionally, we propose a novel probability-based exploration strategy, utilizing a multi-value frontier scoring to enhance lifelong exploration efficiency. Extensive experiments demonstrate the efficiency and robustness of the proposed system.
\end{abstract}


\section{Introduction}
\label{sec:Introduction}
 As a fundamental requirement, ObjectNav is defined as navigating an unseen environment to find a target object\cite{define}, serving as the key to achieving embodied intelligence for robots handling complex household and service tasks\cite{eobodied}. However, the majority of previous research typically focuses on the isolated single object navigation \cite{survey1} \cite{survey2}, overlooking the lifelong memory applicability and hampering effective continual target object navigation. For example, in practical robotic scenarios, navigation rarely involves pursuing an isolated objective (e.g. to locate a \textit{cup}), but rather requires sequentially achieving lifelong open-vocabulary goals as Fig. \ref{first} (e.g., In hotel involving different unseen room scenes, user may first instruct a robot to find a \textit{cup}, and after a while, he desires to find \textit{remote control}). Consequently, how to achieve open-vocabulary lifelong navigation in large scenes remains an urgent problem for mobile service robots.

 Accomplishing open-vocabulary lifelong ObjectNav presents two key challenges: 1) The unpredictable and disordered nature of automated open-vocabulary labeling\cite{open} poses barriers to building structured memory, making it hard to simultaneously achieve open-semantic and lifelong memory representation. In detail, under open-vocabulary autolabel settings, directly storing raw, unpredictable textual labels in memory is impractical—this approach leads to severe efficiency degradation(e.g., repeated storage, synonym representation, unpredictable misrecognition). Previous methods like GOAT\cite{goat}, which employ Mask R-CNN as an open-vocabulary detector, require predefined vocabularies, leading to the inability to retain information about previously encountered objects when new categories are manually added. 
 
 2) The gap between scene information and frontier selection \cite{vlm1} makes it difficult to fully utilize contextual information for efficient reasoning and sustained exploration. 

To mitigate these issues, two critical insights need to be considered: memory maintenance and efficient exploration. First, an effective memory mechanism should focus on capturing the distinctive features of objects rather than directly storing their raw labels. This involves extracting and storing representative features of objects, then using feature matching to determine whether an object is new instance. By replacing error-prone text labels with more robust visual features as the indexing key, the system gains improved resistance to interference and higher retrieval efficiency. Second, inspired by human exploration in unfamiliar settings, the process should not only integrate relevant semantic information, but also prevent repeated visits to the same locations over prolonged exploration.

Through a detailed analysis of the above method, this paper proposes OVAL, a novel lifelong open-vocabulary memory framework capable of achieving structured long-term retention of automated open-vocabulary labeling information. One of the key elements to improve the robustness of the memory model lies in establishing memory descriptors, thereby enabling systematic management of the memory model. By preprocessing and postprocessing automated open-vocabulary labeling terms, the memory model can become more structured and precise for storage and query. To further improve the lifelong exploration effectiveness, we design a probability-based exploration
strategy that integrates distance, positional, and semantic information to improve generalization and interpretability.

\begin{figure*}[t]
    \center
    \includegraphics[width=17.5cm]{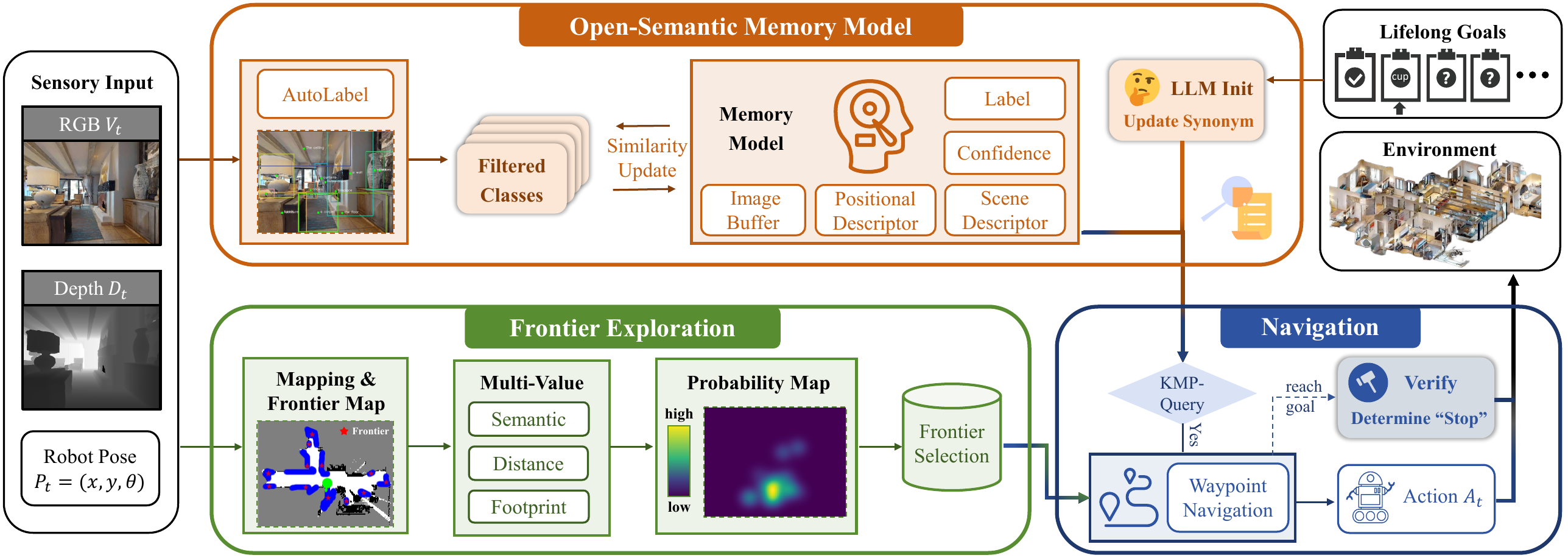}
    \caption{ \textbf{Pipeline of our OVAL:} The frontier exploration module utilizes depth and pose to build a grid map, followed by frontier selection based on probability map. Then, the open-semantic memory model passes through the autolabel extraction, keyword filter, and similarity calculation to update our proposed memory model. Finally, the navigation module searches for goal synonyms via KMP-queries and navigates waypoints, verifying when to stop upon reaching the goal.}
    \label{Fig:framework}
    \vspace{-0.3cm}
\end{figure*}

\subsection{Related Work}
\label{sec:citations}
\subsubsection{Object Goal Navigation}
Existing object navigation methods fall largely into two categories: end-to-end learning and modular methods. End-to-end methods train visual encoders with reinforcement learning (RL) \cite{rl1}\cite{rl2} or imitation learning (IL) \cite{il1}\cite{il2}, mapping observations to actions directly. However, due to the sim-to-real gap, the observations in simulation environments are very likely to be different from the real world. To address these limitations, researchers also explore modular methods \cite{module1}, which usually consist of a mapping module \cite{mapping1}, a policy module \cite{policy2}, and a path planning module \cite{path}. 

With the advancements in multi-modal representation learning, ObjectNav now enables open-vocabulary object localization. While recent studies \cite{llm1}\cite{llm2}\cite{llm4}\cite{llm5} integrate common sense knowledge from large language models (LLMs, e.g.GPT-4 and GPT-4V)  to enable zero-shot common sense reasoning, excessive reliance on an LLM requires a large amount of compute resource. Compared to LLM-based methods, some works \cite{vlm1}\cite{vlm2} implement lightweight vision language models (VLMs, e.g. CLIP and BLIP) executable on standard laptops, producing semantic value maps directly from RGB inputs paired with textual prompts, thereby eliminating the need for intermediate text generation from visual data.Building upon these visual-semantic representations, BeliefMapNav \cite{beliefmapnav} leverages 3D voxel belief maps to integrate LLM reasoning with visual embeddings.  However, these methods frequently struggle to iteratively leverage the corresponding value maps during continual goal-object task exploration, overlooking the functional contributions of memory mechanisms.
\subsubsection{Memory Guided Navigation}
Memory serves as a critical component in enabling intelligent agents to perform spatial reasoning and geometric analysis within 3D environments \cite{memory}. Multi-object navigation (MultiON) \cite{multion, m1, m2, m4} incorporates memory mechanisms but fails to constitute a lifelong framework due to pre-defined task lists. In contrast, GOAT \cite{goat}, OneMap \cite{onemap}, and OVAMOS \cite{ovamos} leverage accumulated experiences for persistent navigation. OneMap \cite{onemap} implements open-vocabulary mapping via feature vectors, whereas the others, alongside 3D-Mem \cite{3d-mem}, remain restricted by predefined category sets.
\subsection{Contribution}
\label{sec:citations}
In summary, our contributions are summarized as follows:
\begin{enumerate}
\item We present a novel explicit memory model that enables the updating and querying of open-semantic memories, facilitating lifelong open-vocabulary ObjectNav.
\item To address reasoning constraints during frontier exploration, we propose a novel probability-based exploration strategy grounded in a multi-value frontier scoring, which significantly enhances lifelong exploration efficiency.
\item Experiments on a challenging interactive simulator validate the capabilities, efficiency, and robustness of our framework on the lifelong open-vocabulary ObjectNav task.
\end{enumerate}

\section{Overview of the Framework}
\label{sec:citations}
\subsection{Problem Statement}  
Traditional ObjectNav requires a robot to locate a goal object $G_t$ (e.g., `cup') without a pre-built map. To enhance practicality, we propose an extended problem where the robot should sequentially search for lifelong open-vocabulary object goals $\hat{G}_t$ (e.g., ``cup $\rightarrow$ remote control$\rightarrow\cdot\cdot\cdot$'') within a previously unseen environment. Unlike MultiON tasks \cite{multion}, lifelong goal navigation assigns subsequent navigation targets conditioned on the completion of the preceding object goal.\par
At each time step, the robot's sensory input comprises an RGB image $V_t$, a corresponding depth image $D_t$, and the current robot pose $P_t$. Based on this observation $O_t=(V_t, D_t, P_t)$, the agent must select an action $A_t$ containing \{Stop, Move Forward, Turn Left, Turn Right, Look Up, Look Down\}. An episode is deemed successful if the STOP action is triggered within 1 meter of any target object and occurs within 500 navigation steps. The process of ObjectNav solving should satisfy: 1) maximizing the number of successful goal explorations, 2) minimizing exploration path length.
\label{sec:citations}
\subsection{System Framework}
\label{sec:citations}
The structure of the proposed OVAL framework completes open-vocabulary lifelong goal navigation tasks through a series of interconnected processes, as shown in Fig. \ref{Fig:framework}. It primarily comprises three modules: the Frontier Exploration'' module (Section \ref{sec:Frontier}), the Open-Semantic Memory Model'' module (Section \ref{sec:Open}), and the ``Navigation'' module (Section \ref{sec:Nav}). The single-episode evaluation of OVAL follows the algorithmic workflow detailed in Algorithm \ref{alg:oval_nav_final}.

\SetKwFor{For}{for}{\string do}{}
\RestyleAlgo{ruled}
\SetKwFor{For}{for}{\string do}{}
\RestyleAlgo{ruled}
\begin{algorithm}[t]
    \caption{\textbf{OVAL Evaluation.}  The algorithm iteratively searches for goal objects in semantic memory, navigates to candidates for verification, and explores frontiers when no valid targets are found.}
    \label{alg:oval_nav_final}
    \LinesNumbered
    \DontPrintSemicolon
    
    \KwIn{Goal $Goal_{original}$, Environment $Env$}
    \KwOut{Success or Failure, SPL}
    
    $Goal \leftarrow \textbf{ProcessGoal}(Goal_{original})$\;
    $M_{sem}, M_{map} \leftarrow \textbf{InitializeMemoryAndMap}(Env)$\;
    $Tried_{goals}, Tried_{frontiers} \leftarrow \emptyset, \emptyset$\;
    $\textbf{LookAround}()$\;

    \While{not $\textbf{EpisodeOver()}$}{
        $Objects_{found} \leftarrow M_{sem}.\textbf{Query}(Goal)$\;
        $Objects_{valid} \leftarrow Objects_{found} \setminus Tried_{goals}$\;
        
        \uIf{$Objects_{valid} \neq \emptyset$}{
            $Object_{target} \leftarrow \textbf{SelectBestCandidate}(Objects_{valid})$\;
            $\textbf{NavigateTo}(Object_{target}.coord)$\;
            $success \leftarrow \textbf{Verify}()$\;
            \uIf{success}{
                $\textbf{Stop}()$\;
                \textbf{break}\;
            }
            \Else{
                $Tried_{goals} \leftarrow Tried_{goals} \cup \{Object_{target}\}$\;
            }
        }
        \Else{
            $Frontiers \leftarrow \textbf{GetFrontiers}(M_{map}) \setminus Tried_{frontiers}$\;
            \uIf{$Frontiers \neq \emptyset$}{
                $Frontier_{chosen} \leftarrow \textbf{ChooseFrontier}(Frontiers, M_{sem})$\;
                $\textbf{NavigateTo}(Frontier_{chosen}.coord)$\;
                $\textbf{LookAround}()$\;
            }
            \Else{
                $\textbf{RandomWalk}()$\;
            }
        }
    }
    \KwRet $\textbf{GetMetrics}()$\;
\end{algorithm}
\SetKwFor{For}{for}{\string do}{}
\RestyleAlgo{ruled}
\SetKwFor{For}{for}{\string do}{}
\RestyleAlgo{ruled}

\section{Implementation}
\label{sec:implementation}
\subsection{Frontier Exploration}
\label{sec:Frontier}
Building upon established methodology\cite{vlm1}, we employ depth and pose to construct a 2D top-down grid map depicting the obstacles encountered by the robot. Specifically, we convert the current depth into a point cloud, exclude points based on the height of the point cloud, transform these points into the global frame, and subsequently project them onto a 2D grid. A frontier refers to the boundary between explored and unexplored regions. To guide exploration, frontier waypoints can be generated by applying DBSCAN clustering to these frontiers as local target points. For frontier selection, we propose a novel probability-based semantic distance exploration scheme. For exploration efficiency and practical frontier accessibility considerations, OVAL records previously attempted frontiers and avoids revisiting them in future exploration, which also prevents oscillation between unreachable frontiers.

The frontier is selected based on a probability, which is determined by three key factors as follows:
\begin{equation}
    \label{equ:exception}
	\begin{aligned}
P(F) \propto o_d(F) + o_s(F) + o_f(F)
\end{aligned}
\end{equation}
The distance factor $o_d(F)$ prioritizes frontiers in closer proximity to accelerate goal-reaching efficiency. The semantic factor $o_s(F)$ integrates room information through environmental semantics - for instance, when locating a toilet, the agent demonstrates a preference for rooms containing characteristic fixtures like sinks, showers, and towel racks. The footprint factor $o_f(F)$ implements exploration diversification by discouraging persistent revisits to previously mapped areas, thereby enforcing systematic coverage across distinct spatial units. Specifically, at time step $t$, the factor values corresponding to each Frontier are defined as follows:
\begin{equation}
    \label{equ:exception}
\begin{cases}
    \begin{aligned}
       o_d(F) &= A_d \mathcal{N}(P_F; P_t, \sigma_d\boldsymbol{I}) \cdot \mathbf{1}_{||P_F-P_t|| \leq d_{th}} \\[1ex]
       o_s(F)&=A_s\sum_{k=0}^{n(\mathcal{C})}\left[\mathcal{N}(P_F; X_k, \sigma_s\boldsymbol{I}) P(G_t,S_k)\right]\\  
       o_f(F)&=-A_f\sum_{k=0}^{t}\mathcal{N}(P_F; P_k, \sigma_f\boldsymbol{I})
  \end{aligned}
\end{cases}
\end{equation}
where $A_d, A_s, A_f$ represent the amplitude magnitudes as weights, $\sigma_d, \sigma_s, \sigma_f$ denote the spatially decaying rates of the corresponding factors, $\boldsymbol{I}$ is the identity matrix, and $\mathcal{N}(x; \boldsymbol{\mu}, \Sigma)$ represents the Gaussian distribution with mean $\mu$ and covariance $\Sigma$ evaluated at $x$.  $d_{th}$ is the distance threshold. $P_{\{\cdot\}}$ represents the agent's position corresponding to the time step ${\{\cdot\}}$, and $P_F$ is the coordinates of the corresponding frontier. $\mathcal{C}$ denotes observed object classes within the memory model, and $X_K, S_K$ represent the corresponding memory descriptors in \ref{memory model}. We pre-defined several groups containing items that frequently co-occur, and also established semantic correlation between items and groups as $P(\cdot,\cdot)$. If the goal does not exist in the predefined table, the model will utilize the other two terms in formula (2) to complete the exploration.
\subsection{Open-Semantic Memory Model}
\label{sec:Open}
\subsubsection{Open-Semantic Preprocess}
\label{Open-Set Semantic}
To enable zero-shot open-vocabulary recognition, we integrate the Autolabel mode of Grounded-SAM2 \cite{groundedsam} into the memory module. However, the Autolabel mode tends to generate substantial noisy and irrelevant textual instances, such as "wall" or "various objects". To mitigate this effect, we implement keyword matching to filter out specific words. Furthermore, to deal with the occurrence of synonymous expressions in autolabel, we leverage LLM at an episode initialization to obtain synonymous classes $\overline{G}_t$ of the goal $G_t$ for completed semantic retrieval. In an open-vocabulary setting, the substantial overlap of semantics precludes the possibility of constructing a coherent semantic map through conventional semantic projection approaches, thereby rendering global query implementation inherently challenging. Accordingly, we propose a memory model in \ref{memory model} that leverages descriptors and confidence to effectively memorize open-vocabulary information.

\subsubsection{Memory Model Definition}
\label{memory model}
The core of OVAL lies in maintaining an open-vocabulary memory model, similar to the human hippocampus during memory consolidation, which orderly captures and stores semantic representations of all observed instances in unknown environments. During lifelong ObjectNav tasks, the memory model efficiently retrieves goals through memory querying and semantic grounding.

In detail, the open-vocabulary memory model comprises various observed object classes, expressed as $\mathcal{C}=\{S_i, \mathcal{I}_i, X_i, H_i, C_i\}_{i=1}^{n(\mathcal{C})}$. Specifically, $S_i$ is the object label generated through the open-semantic preprocessing in Section~\ref{Open-Set Semantic}. The descriptor $\mathcal{C}$ is maintained during each memory update to record the features of the object and facilitate the distinction of instances. $\mathcal{I}_i=\{V_j\}_{j=1}^N$ is the image buffer aligned with the observed object, with a maximum limit of $N$. To distinguish between different instances, $X_i$ serves as a positional descriptor for the corresponding object label, representing the coordinate on the 3D plane mapped from the bounding box's pixel center location $\overline{p}_c$.  For different objects that are close in position (e.g., objects on either side of a wall), $H_i \in \mathbb{R}^{H \times S \times V}$ serves as a scene descriptor by leveraging distinct room color characteristics, which is derived from the HSV color histogram\cite{hsv} representation of the scene.

When observed objects appear image margin or occupy minimal frame space, semantic extraction tends to lead to recognition errors. Object detection and recognition performance is generally enhanced by three key factors: (1) reduced distance between the object and camera sensor (within optimal range), (2) central positioning within the field of view, and (3) increased spatial coverage of the object within the image frame. To mitigate this limitation, the memory model incorporates a confidence probability denoted as $C_i$ for determining stop, which is formally defined as follows:
\begin{equation}
    \label{equ:exception}
	\begin{aligned}
  C_i = \exp({-\sigma\frac{||\overline{p}_c-p_{c}||D_{t,\overline{p}_c}}{A_b}})
\end{aligned}
\end{equation}
where $p_{c}$ denotes the central coordinate of the image pixel, $A_b$ represents the area of the detected object's bounding box. $\sigma$ serves as a hyperparameter that governs the spatial extent of regulation.
\subsubsection{Memory Model Management}

\begin{figure}[t]
    \centering
    \includegraphics[width=9cm]{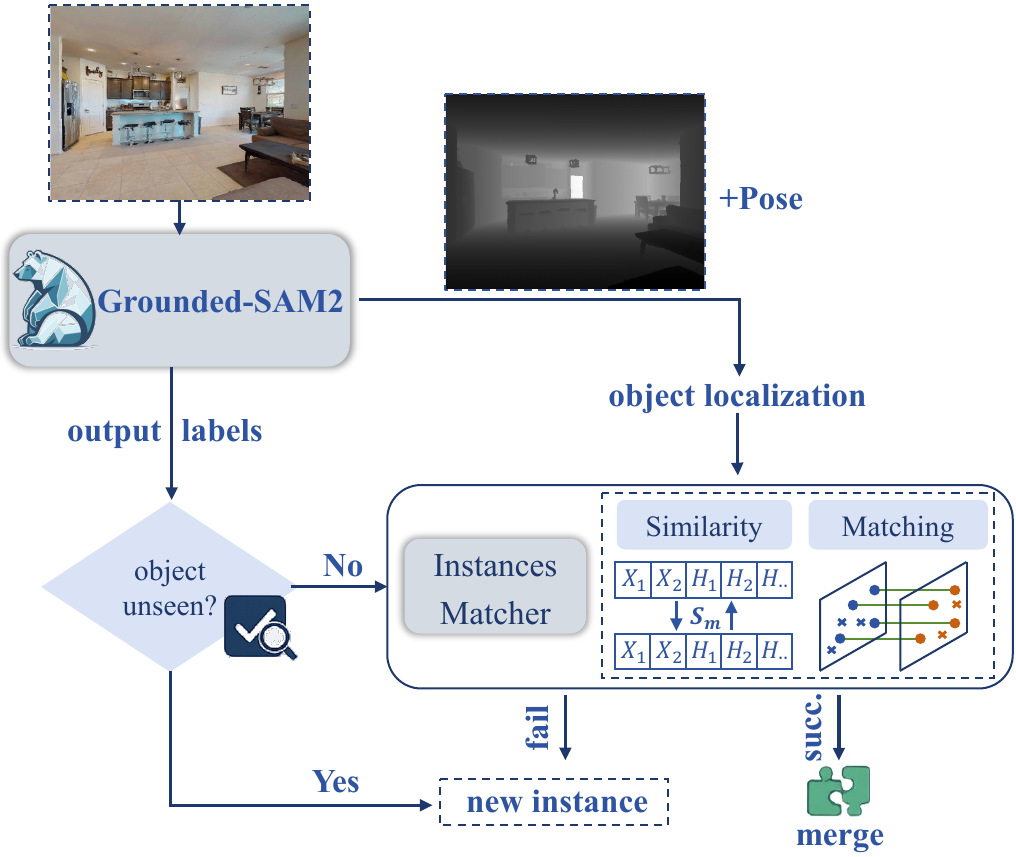}
    \caption{\textbf{The workflow of Memory Model Management.} The system designs the instances matcher module for objects labeled in memory, calculating similarity and matching to identify new instances (Instance refer to different occurrences of the same object category), thereby preventing interference with previously stored memories. }
    \label{mem}
    \vspace{-0.2cm}
\end{figure}

The primary challenge in memory systems lies in mitigating memory interference, which requires the systematic updating of stored information. Although not merging identical instances preserves recognition accuracy, it introduces significant computational overhead, excessive storage requirements (e.g., redundant storage of 'bed' across thousands of instances), and substantially impairs query efficiency. Consequently, instance merging is crucial for real-time performance optimization. Specifically, when encountering a newly observed object class $\{S_j, \mathcal{I}_j, X_j, H_j, C_j\}$, the system must perform a rapid judgment to determine whether the object represents a novel instance or corresponds to a pre-existing entry $\{S_i, \mathcal{I}_i, X_i, H_i, C_i\}$ within the memory framework, thereby enabling stable memory update.  In detail, as Fig. \ref{mem}, when the label $S_j$ does not exist in $\mathcal{C}$, it must indicate a newly emerged object without memory; however, for the label $S_j$ that does exist in $\mathcal{C}$, further judgment is required to determine whether they are instance from memory. To streamline screening and eliminate clearly inconsistent scenarios, we initially compute the similarity using scene descriptors and positional descriptors, as illustrated below:
\begin{equation}
    \label{equ:exception}
	\begin{aligned}
S_m = \lambda_{H} \text{Sim}\left(H_i, H_j\right)
- \lambda_{X} 	
\text{Sigmoid}(k||X_i-X_j||)
\end{aligned}
\end{equation}
where $\lambda_{H}$, $\lambda_{X}$ denote the similarity weights of the scene descriptor and positional descriptor, respectively. The $\text{Sim}(\cdot,\cdot)$ function computes the element-wise minimum across HSV channels. For most newly observed objects exhibiting either relatively low or exceptionally high similarity, direct determination can be made through metric $S_m$. In scenarios demonstrating ambiguous similarity $(\tau_{l}\leq{S}_{m}\leq\tau_{u}$, $\tau_{u}/\tau_{l}$ are upper/lower thresholds derived from descriptor similarity), we deploy SuperGlue \cite{superglue} to match instances whose $|\mathcal{M}_{s g}|\geq\tau_{M}$ ($\tau_{M}$ is the minimum matching points threshold, $|\cdot|$ counts the number of elements in a set).
\begin{subequations}
    \label{equ:exception}
    \begin{align}
\mathcal{M}_{sg}=\{(p_{m},q_{n})\mid 
&\text{score}(p_{m},q_{n})>\tau_{sg}\}\\
\text{where} \quad p_{m}\in{\mathcal{F}}&(\mathcal{I}_i),\;q_{n}\in{\mathcal{F}}(\mathcal{I}_j)
\end{align}
\end{subequations}
 where $\mathcal{F}(\cdot)$ denotes the local feature points extracted from the image by SuperGlue, $\text{score}(\cdot, \cdot)$ computes the similarity score between feature point pairs, and $\tau_{sg}$ controls the quality of matching.

When the match fails, the memory model instantiates a new object class to model the emergent instance. If the match succeeds and confirms the instance originates from existing memory, parameter updates are governed by the confidence probability to merge instances: If $C_i\le C_j$, the original object class parameters $X_i, H_i, C_i$ will be systematically overwritten by $X_j, H_j, C_j$, with $\mathcal{I}_i$ refined via $\mathcal{I}_j$. 

\subsection{Navigation}
\label{sec:Nav}
For the lexical heterogeneity issue in open-domain semantic scenarios, such as encountering memory model labels containing ``green desk" while querying ``desk", we employ the KMP substring search algorithm to resolve cross-granularity querying challenges inherent in open semantic environments. When a goal object $G_t$ and its synonymous classes $\overline{G}_t$ are queried in the memory model and indexed as $\mathcal{C}_i$, the robot's trajectory is computed via the greedy path planner\cite{path2} toward the PointNav target $X_i$. To prevent peripheral objects, OVAL performs a verification operation after reaching $X_i$, checking via an LLM whether it has arrived at the final target. Firstly, it conducts a 360-degree panoramic capture around the goal to obtain a set of $K$ panoramic images $\{V_k\}_{k=1}^K$. Then, it uses the LLM to obtain the probability that the goal is present in the image set and combines this with the confidence probability from the Memory Model through weighted scoring, as follows:
 \begin{equation}
    \label{equ:exception}
	\begin{aligned}
S = \omega \phi(\{V_k\}_{k=1}^{K})+(1-\omega)C_i
\end{aligned}
\end{equation}
where $\phi(\cdot)$ represents the probability output by the LLM, and $\omega$ denotes the confidence weight assigned to the LLM. If score $S$ exceeds the threshold, the agent outputs 'STOP'. If the $S$ fails to meet the threshold, the corresponding instance class is removed from the memory model. It is observed that the model exhibits low sensitivity to the precise value of this threshold, as its primary function is to effectively filter out instances with exceptionally low scores, thereby significantly enhancing overall model performance by reducing statistical noise.

\section{EXPERIMENTS}
\subsection{Experiment Setup }

\textbf{Implementation Details}:
We utilize GPT4o to verify whether the goal is reached and GPT4o-mini to generate synonyms. In the Grounded-SAM2 \cite{groundedsam} module, we employ the $<$ DETAILED\_CAPTION $>$ setting and leverage Florence-2-base for further processing. Our experiments are conducted on a machine equipped with the RTX 4090D (24GB) hardware configuration. The key parameters are set as follows: $A_d=1, A_s= 0.5,Af=1,\Sigma_d=1e6,\Sigma_s=1e6,\Sigma_f=1e6,\lambda_H=0.5,\lambda_X=0.5,\tau_{l}=0.2,\tau_{u}=0.8,\tau_{M}=60,\omega=0.5,k=8$.

\textbf{Datasets and Metric}: We evaluate our approach using the Habitat \cite{habitat} simulator on the validation splits of two different datasets: HM3D \cite{hm3d} and MP3D \cite{mp3d}. Following the experimental setup of proven open-vocabulary navigation methods ESC \cite{llm2}, VLFM \cite{vlm1}, and PixNav \cite{llm5 }, we use the same datasets to demonstrate our method's effectiveness in open-vocabulary scenarios. We adopt Success Rate (SR) and Success Rate Weighted by Inverse Path Length (SPL) as the evaluation metrics \cite{define}, which is a metric evaluating the navigation path efficiency.
\subsection{Lifelong ObjectNav Dataset Generation Method}
\label{subsec:dataset_generation}

All Lifelong ObjectNav datasets are derived from standard ObjectNav datasets (e.g., HM3D, MP3D) through strategic episode reordering and state persistence mechanisms.

\textbf{Episode Sorting} To ensure temporal locality for tasks within the same spatial area, episodes are reordered as follows:

\begin{enumerate}
    \item Scene ID: Episodes are first grouped by scene.
    \item Floor Height: Within each scene, episodes are further grouped by floor, where positions with height difference less than $0.5$m are considered the same floor.
\end{enumerate}

This sorting creates episode groups where each group contains multiple navigation tasks within the same scene and floor area.

\textbf{Lifelong State Persistence Logic} At each episode transition:

\begin{itemize}
    \item Same Group Continuation: If the new episode belongs to the same group (same scene ID and floor), the agent retains its previous map and memory while teleporting to the previous episode's end position.
    
    \item Group Switch: When transitioning to a different group, standard ObjectNav initialization occurs (clear map/memory, use new episode's start pose).
\end{itemize}

This approach enables lifelong learning by allowing agents to accumulate spatial knowledge within each group while naturally resetting when moving to entirely different environments.

\subsection{Evaluation of Lifelong ObjectNav }

 \begin{table*}[t]
\caption{\textbf{Comparison with SOTA methods on HM3D and MP3D datasets (1000 episodes) for lifelong ObjectNav. }The best results are in \textbf{bold}. N/A denotes a run failure in the dataset.}
\centering
\resizebox{0.8\textwidth}{!}{
        \large
        
\begin{tabular}{c|cc||cc|cc}
\toprule
  \multirow{2}{*}{\textbf{Method}}   & \multirow{2}{*}{\textbf{Open-Vocabulary}} & \multirow{2}{*}{\textbf{Lifelong Memory}}&\multicolumn{2}{c|}{\textbf{HM3D}}&\multicolumn{2}{c}{\textbf{MP3D}}\\
   &&&SR($\%$)$\uparrow$ & SPL($\%$)$\uparrow$ & SR($\%$)$\uparrow$& SPL($\%$)$\uparrow$\\
   \midrule
  VLFM  \cite{vlm1} & $\checkmark$ & $\times$ &53.7&30.6&35.1&16.7\\
   GOAT \cite{goat} & $\times$ & $\checkmark$ &59.2&31.2&N/A&N/A\\
  OVAL (Ours)& $\checkmark$ & $\checkmark$ &\textbf{68.1}&\textbf{33.8}&\textbf{44.1}&\textbf{18.6}\\
\bottomrule
\end{tabular}
}

    \vspace{-1em}
\label{lifelong}
\end{table*}

\begin{figure}[t]
    \centering
    \includegraphics[width=7.5cm]{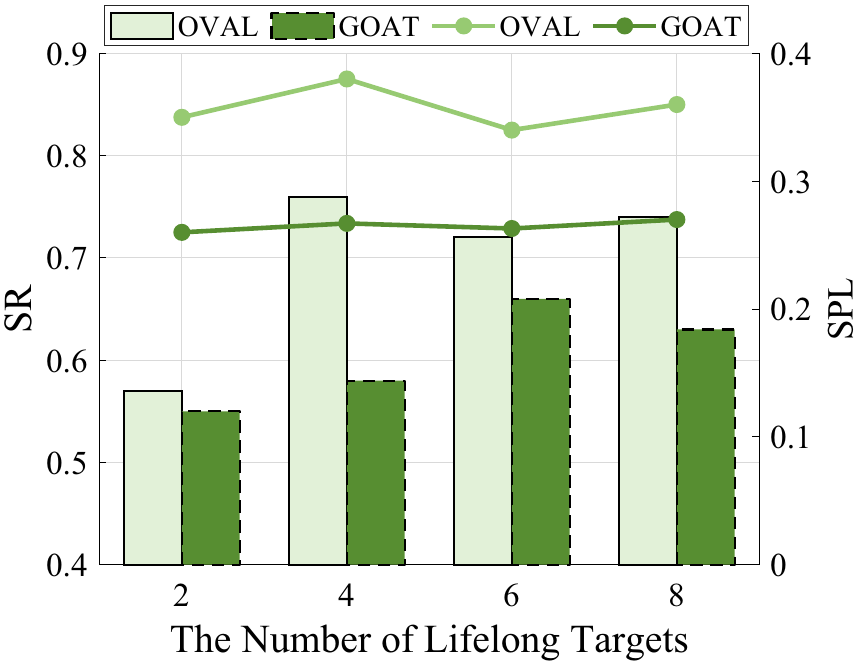}
    \caption{\textbf{Lifelong ObjectNav Tasks Performance Comparison with GOAT in HM3D.} Bar and line charts show SR and SPL comparisons, respectively.}
    \label{com}
    \vspace{-0.2cm}
\end{figure}

\begin{figure}[t]
    \centering
    \vspace{0.32cm}
    \includegraphics[width=7.5cm]{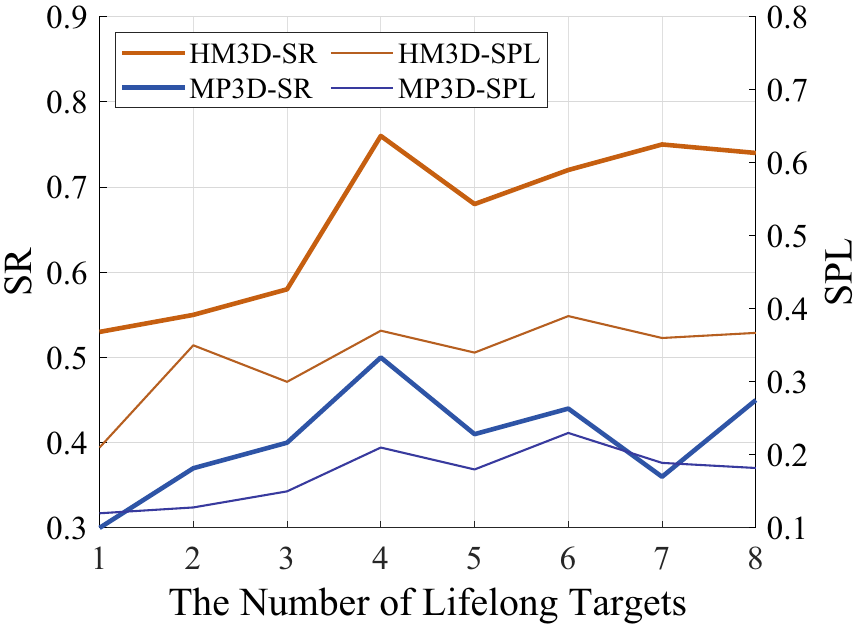}
    \caption{Ablation study on the performance of varying numbers of lifelong targets in the Lifelong Object Nav Tasks. }
    \label{ablition}
    \vspace{-0.2cm}
\end{figure}

\begin{table}[t]
\caption{Ablation study on verify ``STOP", memory model, and probability map exploration strategy method on HM3D datasets.}
\centering
\resizebox{0.48\textwidth}{!}{
        \large
        
  \begin{tabular}{c|c|c|cc}  
    \toprule
    Verify ``STOP"& Memory Model& Probability Map & SR($\%$)$\uparrow$ & SPL($\%$) $\uparrow$ \\  
    \midrule
  $\times$ & \checkmark & \checkmark  & 61.3 & 31.2\\
    \checkmark &$\times$ & \checkmark  & 56.1 & 26.0  \\  
    \checkmark &\checkmark & $\times$& 66.8 & 32.3  \\  
    \checkmark &\checkmark & \checkmark & \textbf{68.1} & \textbf{33.8} \\  
    \bottomrule
   
  \end{tabular}
}

    \vspace{-1em}
\label{abl}

\end{table}
To evaluate the performance of the lifelong ObjectNav task, we compare our methods with the SOTA classical ObjectNav method VLFM \cite{vlm1}, and memory-inclusive method GOAT \cite{goat}. Regarding real-time operational efficiency, OVAL exhibits a Frames Per Second (FPS) rate of 1.14, which is on a comparable order of magnitude to VLFM (1.67 FPS) and GOAT (1.85 FPS).Including LLM overhead, 1 FPS is largely sufficient for real-time operation via asynchronous execution. 

As illustrated in Table \ref{lifelong}, our proposed method OVAL attains superior SR and SPL results compared to other SOTA ObjectNav approaches in HM3D and MP3D. VLFM lacks memory capabilities, resulting in similar success rates between lifelong ObjectNav and standard ObjectNav tasks. The GOAT recognition method cannot process open-vocabulary objects that it hasn't been trained on, leading to its failure in the MP3D dataset containing complex and diverse object categories. In addition, as shown in the Fig. \ref{com}, we compared the performance of OVAL and GOAT under different numbers of lifelong targets. On the one hand, our method of storing objects using coordinates and confidence is more concise, intuitive, and easier to maintain compared to the traditional approach based on point cloud projection. On the other hand, the confidence system enables us to have a higher probability of selecting the correct goal while not neglecting the exploration of potential targets, thus achieving a better balance between precision and recall.

We conduct ablation experiments for the lifelong ObjectNav on the HM3D dataset for our proposed method in Table \ref{abl}. The verify ``STOP'' module significantly improves the SR of single ObjectNav tasks without affecting the SPL metric. The memory system is the core of the lifelong  ObjectNav problem and greatly enhances the SR and SPL of the navigation. The probability map improves navigation performance in the first few navigations, and then its advantage gradually diminishes as the memory system is gradually refined. In addition, as shown in Fig. \ref{ablition}, we conduct ablation experiments on varying numbers of target objects within lifelong scenarios. The experiment demonstrate that as the number of targets increases, SR and SPL exhibit significant improvement, confirming the effectiveness of our memory model in the lifelong ObjectNav task. When the number of targets reaches $4$ or above, the memory model tends to stabilize, leading to a performance boost.

\subsection{Evaluation of ObjectNav }

As Table. \ref{all}, we evaluate our method on the traditional ObjectNav task and compare it with state-of-the-art object navigation models. On the HM3D dataset, our method achieves state-of-the-art success rate among training-free models. On the MP3D dataset, our method attains the highest success rate overall,  demonstrating its adaptability to various types of goals. OVAL demonstrates superior performance on ObjectNav tasks due to two key advantages: the verification mechanism effectively leverages panoramic images to accurately filter out VLM identification errors, and the enhanced exploration strategy not only suppresses redundant exploration tendencies but also incorporates semantic considerations to guide navigation decisions.

To analyze the impact of various elements in the Probability Map on the exploration process, we conduct targeted ablation studies on different components including footprint, distance, and semantic information, as demonstrated in Table \ref{tab:probability_map_ablation}.
The footprint component proves most crucial by preventing redundant exploration. The semantic component enhances performance by enabling efficient exploration through semantic associations with common objects, while the distance component improves efficiency by prioritizing exploration in nearby areas.

\begin{table*}[t]
\caption{\textbf{Comparison with SOTA methods on HM3D and MP3D datasets (1000 episodes) for object goal navigation task.  } The best results and second best are in \textbf{bold} and \underline{underlined}, respectively. $-$ represents that the corresponding data is not in the original report.}
\centering
\resizebox{0.9\textwidth}{!}{
        \large
        
\begin{tabular}{c|ccc||cc|cc}
\toprule
  \multirow{2}{*}{\textbf{Method}}  & \multirow{2}{*}{\textbf{Open-Vocabulary}} & \multirow{2}{*}{\textbf{Training Free}}&\multirow{2}{*}{\textbf{Lifelong Memory}} &\multicolumn{2}{c|}{\textbf{HM3D}}&\multicolumn{2}{c}{\textbf{MP3D}}\\
   &&&&SR($\%$)$\uparrow$ & SPL($\%$)$\uparrow$ & SR($\%$)$\uparrow$& SPL($\%$)$\uparrow$\\
   \midrule
   PONI \cite{poni}& $\times$ & $\times$ & $\times$ &-&-&31.8&12.1\\
    SemEXP \cite{module1}& $\times$ & $\times$ & $\times$ &-&-&36.0&14.4\\

           L3MVN \cite{llm4} & $\times$ & $\checkmark$ & $\times$ &50.4&23.1&-&-\\
  \midrule
      Habitat-Web\cite{il1}& $\checkmark$ & $\times$& $\times$ &41.5&16.0&31.6&8.5\\
      ZSON \cite{zson}& $\checkmark$ & $\times$ & $\times$&25.5&12.6&15.3&4.8\\
 OVRL \cite{ovrl}& $\checkmark$ & $\times$& $\times$ &-&-&28.6&7.4\\
  OVRL-V2 \cite{ovrlv2}& $\checkmark$ & $\times$& $\times$ &64.7&28.1&-&-\\
    PixNav \cite{llm5} & $\checkmark$ & $\times$& $\times$ &37.9&20.5&-&-\\

  \midrule
     ESC \cite{llm2}& $\checkmark$ & $\checkmark$& $\times$ &39.2&22.3&28.7&14.2\\
    VLFM \cite{vlm1} & $\checkmark$ & $\checkmark$& $\times$  &52.5&\textbf{30.4}&\underline{36.4}&\textbf{17.5}\\
 VoroNav \cite{voronav}& $\checkmark$ & $\checkmark$& $\times$ &42.0&26.0&-&-\\
 OpenFMNav \cite{openfmnav} & $\checkmark$ & $\checkmark$ & $\times$&54.9&24.4&-&-\\
 TopV-Nav \cite{topv}& $\checkmark$ & $\checkmark$ & $\times$&45.9&\underline{28.0}&-&-\\
 Instruct-Nav \cite{instructnav}& $\checkmark$ & $\checkmark$ & $\times$&\underline{58.0}&20.9&-&-\\
   \midrule
       GOAT \cite{goat}& $\times$ & $\checkmark$& $\checkmark$ &50.6&24.1&-&-\\
  OVAL (Ours)& $\checkmark$ & $\checkmark$& $\checkmark$ &\textbf{58.2}&24.5&\textbf{41.1}&\underline{15.3}\\
\bottomrule
\end{tabular}
}

    \vspace{-0.2cm}
\label{all}
\end{table*}
\begin{table}[t]
  \caption{Ablation study on probability map on HM3D ObjectNav.}
\centering
\resizebox{0.48\textwidth}{!}{
        \large
  \begin{tabular}{c|c|c|cc}  
    \toprule
    $o_f(F)$ footprint & $o_d(F)$ distance& $o_s(F)$ semantics& SR($\%$)$\uparrow$ & SPL($\%$) $\uparrow$ \\  
    \midrule
  $\times$ & \checkmark & \checkmark  & 54.3 & 21.4\\
    \checkmark &$\times$ & \checkmark  & 56.7 & 22.8  \\  
    \checkmark &\checkmark & $\times$& 56.2 & 23.9  \\  
    \checkmark &\checkmark & \checkmark & \textbf{58.2} & \textbf{24.5} \\  
    \bottomrule
   
  \end{tabular}
}
    \vspace{-1em}
\label{tab:probability_map_ablation}
\end{table}

\section{Conclusion and Future Work}
\label{sec:conclusion}
This study presents OVAL, a lifelong open-vocabulary memory framework to achieve continual objects goal navigation. To maintain the memory order, we introduce corresponding approach of establishing memory descriptors. A novel probability-based exploration strategy based on multi-value frontier scoring, has been proposed to enhance exploration efficiency. Extensive experiments on the challenging interactive simulator demonstrate the generalization and effectiveness of our reusable memory method. 

\textbf{Limitations}: The current OVAL system still only supports ObjectNav and cannot accommodate inputs from multiple modalities. Additionally, the accuracy of recognition will be affected in dynamic, complex, and fast-moving scenarios. In the future, we will design models that can handle various forms of input and develop more robust recognition and memory systems to accomplish the lifelong navigation task in complex multi-modal scenarios.

{
    \balance
    \bibliographystyle{IEEEtran}
    \bibliography{main}

@article{beliefmapnav,
  title={Beliefmapnav: 3d voxel-based belief map for zero-shot object navigation},
  author={Zhou, Zibo and Hu, Yue and Zhang, Lingkai and Li, Zonglin and Chen, Siheng},
  journal={arXiv preprint arXiv:2506.06487},
  year={2025}
}

@inproceedings{onemap,
  title={One map to find them all: Real-time open-vocabulary mapping for zero-shot multi-object navigation},
  author={Busch, Finn Lukas and Homberger, Timon and Ortega-Peimbert, Jes{\'u}s and Yang, Quantao and Andersson, Olov},
  booktitle={2025 IEEE International Conference on Robotics and Automation (ICRA)},
  pages={14835--14842},
  year={2025},
  organization={IEEE}
}

@article{ovamos,
  title={OVAMOS: A Framework for Open-Vocabulary Multi-Object Search in Unknown Environments},
  author={Wang, Qianwei and Xu, Yifan and Kamat, Vineet and Menassa, Carol},
  journal={arXiv preprint arXiv:2503.02106},
  year={2025}
}

@inproceedings{3d-mem,
  title={3D-mem: 3D scene memory for embodied exploration and reasoning},
  author={Yang, Yuncong and Yang, Han and Zhou, Jiachen and Chen, Peihao and Zhang, Hongxin and Du, Yilun and Gan, Chuang},
  booktitle={Proceedings of the Computer Vision and Pattern Recognition Conference},
  pages={17294--17303},
  year={2025}
}

@inproceedings{hsv,
  title={Segmentation and histogram generation using the HSV color space for image retrieval},
  author={Sural, Shamik and Qian, Gang and Pramanik, Sakti},
  booktitle={Proceedings. international conference on image processing},
  volume={2},
  pages={II--II},
  year={2002},
  organization={IEEE}
}

@article{define,
  title={On evaluation of embodied navigation agents},
  author={Anderson, Peter and Chang, Angel and Chaplot, Devendra Singh and Dosovitskiy, Alexey and Gupta, Saurabh and Koltun, Vladlen and Kosecka, Jana and Malik, Jitendra and Mottaghi, Roozbeh and Savva, Manolis and others},
  journal={arXiv preprint arXiv:1807.06757},
  year={2018}
}

@article{survey1,
  title={A survey of object goal navigation},
  author={Sun, Jingwen and Wu, Jing and Ji, Ze and Lai, Yu-Kun},
  journal={IEEE Transactions on Automation Science and Engineering},
  year={2024},
  publisher={IEEE}
}

@inproceedings{eobodied,
  title={Object goal navigation in eobodied ai: A survey},
  author={Li, Baosheng and Han, Jishui and Cheng, Yuan and Tan, Chong and Qi, Peng and Zhang, Jianping and Li, Xiaolei},
  booktitle={Proceedings of the 2022 4th International Conference on Video, Signal and Image Processing},
  pages={87--92},
  year={2022}
}

@inproceedings{survey2,
  title={A survey of object goal navigation: Datasets, metrics and methods},
  author={Wang, Dewei and Chen, Jiaming and Cheng, Jiyu},
  booktitle={2023 IEEE International Conference on Mechatronics and Automation (ICMA)},
  pages={2171--2176},
  year={2023},
  organization={IEEE}
}

@inproceedings{rl1,
  title={Auxiliary tasks and exploration enable objectgoal navigation},
  author={Ye, Joel and Batra, Dhruv and Das, Abhishek and Wijmans, Erik},
  booktitle={Proceedings of the IEEE/CVF international conference on computer vision},
  pages={16117--16126},
  year={2021}
}

@inproceedings{rl2,
  title={Search for or navigate to? dual adaptive thinking for object navigation},
  author={Dang, Ronghao and Wang, Liuyi and He, Zongtao and Su, Shuai and Tang, Jiagui and Liu, Chengju and Chen, Qijun},
  booktitle={Proceedings of the IEEE/CVF International Conference on Computer Vision},
  pages={8250--8259},
  year={2023}
}

@inproceedings{il1,
  title={Habitat-web: Learning embodied object-search strategies from human demonstrations at scale},
  author={Ramrakhya, Ram and Undersander, Eric and Batra, Dhruv and Das, Abhishek},
  booktitle={Proceedings of the IEEE/CVF conference on computer vision and pattern recognition},
  pages={5173--5183},
  year={2022}
}

@inproceedings{il2,
  title={Pirlnav: Pretraining with imitation and rl finetuning for objectnav},
  author={Ramrakhya, Ram and Batra, Dhruv and Wijmans, Erik and Das, Abhishek},
  booktitle={Proceedings of the IEEE/CVF Conference on Computer Vision and Pattern Recognition},
  pages={17896--17906},
  year={2023}
}

@article{module1,
  title={Object goal navigation using goal-oriented semantic exploration},
  author={Chaplot, Devendra Singh and Gandhi, Dhiraj Prakashchand and Gupta, Abhinav and Salakhutdinov, Russ R},
  journal={Advances in Neural Information Processing Systems},
  volume={33},
  pages={4247--4258},
  year={2020}
}

@inproceedings{habitat,
  title={Habitat: A platform for embodied ai research},
  author={Savva, Manolis and Kadian, Abhishek and Maksymets, Oleksandr and Zhao, Yili and Wijmans, Erik and Jain, Bhavana and Straub, Julian and Liu, Jia and Koltun, Vladlen and Malik, Jitendra and others},
  booktitle={Proceedings of the IEEE/CVF international conference on computer vision},
  pages={9339--9347},
  year={2019}
}

@article{hm3d,
  title={Habitat-matterport 3d dataset (hm3d): 1000 large-scale 3d environments for embodied ai},
  author={Ramakrishnan, Santhosh K and Gokaslan, Aaron and Wijmans, Erik and Maksymets, Oleksandr and Clegg, Alex and Turner, John and Undersander, Eric and Galuba, Wojciech and Westbury, Andrew and Chang, Angel X and others},
  journal={arXiv preprint arXiv:2109.08238},
  year={2021}
}

@article{mp3d,
  title={Matterport3d: Learning from rgb-d data in indoor environments},
  author={Chang, Angel and Dai, Angela and Funkhouser, Thomas and Halber, Maciej and Niessner, Matthias and Savva, Manolis and Song, Shuran and Zeng, Andy and Zhang, Yinda},
  journal={arXiv preprint arXiv:1709.06158},
  year={2017}
}

@inproceedings{memory,
  title={Multi-Object Navigation in real environments using hybrid policies},
  author={Sadek, Assem and Bono, Guillaume and Chidlovskii, Boris and Baskurt, Atilla and Wolf, Christian},
  booktitle={2023 IEEE International Conference on Robotics and Automation (ICRA)},
  pages={4085--4091},
  year={2023},
  organization={IEEE}
}

@article{multion,
  title={Multion: Benchmarking semantic map memory using multi-object navigation},
  author={Wani, Saim and Patel, Shivansh and Jain, Unnat and Chang, Angel and Savva, Manolis},
  journal={Advances in Neural Information Processing Systems},
  volume={33},
  pages={9700--9712},
  year={2020}
}

@article{m1,
  title={OVAMOS: A Framework for Open-Vocabulary Multi-Object Search in Unknown Environments},
  author={Wang, Qianwei and Xu, Yifan and Kamat, Vineet and Menassa, Carol},
  journal={arXiv preprint arXiv:2503.02106},
  year={2025}
}

@inproceedings{m2,
  title={Multi-Object Navigation with dynamically learned neural implicit representations},
  author={Marza, Pierre and Matignon, Laetitia and Simonin, Olivier and Wolf, Christian},
  booktitle={Proceedings of the IEEE/CVF International Conference on Computer Vision},
  pages={11004--11015},
  year={2023}
}

@article{m4,
  title={Find Everything: A General Vision Language Model Approach to Multi-Object Search},
  author={Choi, Daniel and Fung, Angus and Wang, Haitong and Tan, Aaron Hao},
  journal={arXiv preprint arXiv:2410.00388},
  year={2024}
}

@inproceedings{goat,
  author       = {Matthew Chang and
                  Th{\'{e}}ophile Gervet and
                  Mukul Khanna and
                  Sriram Yenamandra and
                  Dhruv Shah and
                  So Yeon Min and
                  Kavit Shah and
                  Chris Paxton and
                  Saurabh Gupta and
                  Dhruv Batra and
                  Roozbeh Mottaghi and
                  Jitendra Malik and
                  Devendra Singh Chaplot},
  title        = {{GOAT:} {GO} to Any Thing},
  booktitle    = {Robotics: Science and Systems XX, Delft, The Netherlands, July 15-19,
                  2024},
  year         = {2024},
}

@article{llm1,
  title={CogNav: Cognitive Process Modeling for Object Goal Navigation with LLMs},
  author={Cao, Yihan and Zhang, Jiazhao and Yu, Zhinan and Liu, Shuzhen and Qin, Zheng and Zou, Qin and Du, Bo and Xu, Kai},
  journal={arXiv preprint arXiv:2412.10439},
  year={2024}
}

@inproceedings{llm2,
  title={Esc: Exploration with soft commonsense constraints for zero-shot object navigation},
  author={Zhou, Kaiwen and Zheng, Kaizhi and Pryor, Connor and Shen, Yilin and Jin, Hongxia and Getoor, Lise and Wang, Xin Eric},
  booktitle={International Conference on Machine Learning},
  pages={42829--42842},
  year={2023},
  organization={PMLR}
}

@inproceedings{llm4,
  title={L3mvn: Leveraging large language models for visual target navigation},
  author={Yu, Bangguo and Kasaei, Hamidreza and Cao, Ming},
  booktitle={2023 IEEE/RSJ International Conference on Intelligent Robots and Systems (IROS)},
  pages={3554--3560},
  year={2023},
  organization={IEEE}
}

@inproceedings{llm5,
  title={Bridging zero-shot object navigation and foundation models through pixel-guided navigation skill},
  author={Cai, Wenzhe and Huang, Siyuan and Cheng, Guangran and Long, Yuxing and Gao, Peng and Sun, Changyin and Dong, Hao},
  booktitle={2024 IEEE International Conference on Robotics and Automation (ICRA)},
  pages={5228--5234},
  year={2024},
  organization={IEEE}
}

@inproceedings{vlm1,
  title={Vlfm: Vision-language frontier maps for zero-shot semantic navigation},
  author={Yokoyama, Naoki and Ha, Sehoon and Batra, Dhruv and Wang, Jiuguang and Bucher, Bernadette},
  booktitle={2024 IEEE International Conference on Robotics and Automation (ICRA)},
  pages={42--48},
  year={2024},
  organization={IEEE}
}

@article{vlm2,
  title={WMNav: Integrating Vision-Language Models into World Models for Object Goal Navigation},
  author={Nie, Dujun and Guo, Xianda and Duan, Yiqun and Zhang, Ruijun and Chen, Long},
  journal={arXiv preprint arXiv:2503.02247},
  year={2025}
}

@article{path,
  title={A fast marching level set method for monotonically advancing fronts.},
  author={Sethian, James A},
  journal={proceedings of the National Academy of Sciences},
  volume={93},
  number={4},
  pages={1591--1595},
  year={1996}
}

@article{policy2,
  title={Hierarchical landmark policy optimization for visual indoor navigation},
  author={Staroverov, Aleksei and Panov, Aleksandr I},
  journal={IEEE Access},
  volume={10},
  pages={70447--70455},
  year={2022},
  publisher={IEEE}
}

@inproceedings{mapping1,
  title={Cognitive mapping and planning for visual navigation},
  author={Gupta, Saurabh and Davidson, James and Levine, Sergey and Sukthankar, Rahul and Malik, Jitendra},
  booktitle={Proceedings of the IEEE conference on computer vision and pattern recognition},
  pages={2616--2625},
  year={2017}
}

@inproceedings{poni,
  title={Poni: Potential functions for objectgoal navigation with interaction-free learning},
  author={Ramakrishnan, Santhosh Kumar and Chaplot, Devendra Singh and Al-Halah, Ziad and Malik, Jitendra and Grauman, Kristen},
  booktitle={Proceedings of the IEEE/CVF Conference on Computer Vision and Pattern Recognition},
  pages={18890--18900},
  year={2022}
}

@inproceedings{ovrl,
  title={Offline visual representation learning for embodied navigation},
  author={Yadav, Karmesh and Ramrakhya, Ram and Majumdar, Arjun and Berges, Vincent-Pierre and Kuhar, Sachit and Batra, Dhruv and Baevski, Alexei and Maksymets, Oleksandr},
  booktitle={Workshop on Reincarnating Reinforcement Learning at ICLR 2023},
  year={2023}
}

@article{ovrlv2,
  title={Ovrl-v2: A simple state-of-art baseline for imagenav and objectnav},
  author={Yadav, Karmesh and Majumdar, Arjun and Ramrakhya, Ram and Yokoyama, Naoki and Baevski, Alexei and Kira, Zsolt and Maksymets, Oleksandr and Batra, Dhruv},
  journal={arXiv preprint arXiv:2303.07798},
  year={2023}
}

@article{zson,
  title={Zson: Zero-shot object-goal navigation using multimodal goal embeddings},
  author={Majumdar, Arjun and Aggarwal, Gunjan and Devnani, Bhavika and Hoffman, Judy and Batra, Dhruv},
  journal={Advances in Neural Information Processing Systems},
  volume={35},
  pages={32340--32352},
  year={2022}
}

@article{voronav,
  title={Voronav: Voronoi-based zero-shot object navigation with large language model},
  author={Wu, Pengying and Mu, Yao and Wu, Bingxian and Hou, Yi and Ma, Ji and Zhang, Shanghang and Liu, Chang},
  journal={arXiv preprint arXiv:2401.02695},
  year={2024}
}

@article{openfmnav,
  title={Openfmnav: Towards open-set zero-shot object navigation via vision-language foundation models},
  author={Kuang, Yuxuan and Lin, Hai and Jiang, Meng},
  journal={arXiv preprint arXiv:2402.10670},
  year={2024}
}

@article{topv,
  title={TopV-Nav: Unlocking the Top-View Spatial Reasoning Potential of MLLM for Zero-shot Object Navigation},
  author={Zhong, Linqing and Gao, Chen and Ding, Zihan and Liao, Yue and Liu, Si},
  journal={arXiv preprint arXiv:2411.16425},
  year={2024}
}

@article{instructnav,
  title={Instructnav: Zero-shot system for generic instruction navigation in unexplored environment},
  author={Long, Yuxing and Cai, Wenzhe and Wang, Hongcheng and Zhan, Guanqi and Dong, Hao},
  journal={arXiv preprint arXiv:2406.04882},
  year={2024}
}

@inproceedings{superglue,
  title={Superglue: Learning feature matching with graph neural networks},
  author={Sarlin, Paul-Edouard and DeTone, Daniel and Malisiewicz, Tomasz and Rabinovich, Andrew},
  booktitle={Proceedings of the IEEE/CVF conference on computer vision and pattern recognition},
  pages={4938--4947},
  year={2020}
}

@article{groundedsam,
  title={Grounded sam: Assembling open-world models for diverse visual tasks},
  author={Ren, Tianhe and Liu, Shilong and Zeng, Ailing and Lin, Jing and Li, Kunchang and Cao, He and Chen, Jiayu and Huang, Xinyu and Chen, Yukang and Yan, Feng and others},
  journal={arXiv preprint arXiv:2401.14159},
  year={2024}
}

@article{path2,
  title={Visual memory for robust path following},
  author={Kumar, Ashish and Gupta, Saurabh and Fouhey, David and Levine, Sergey and Malik, Jitendra},
  journal={Advances in neural information processing systems},
  volume={31},
  year={2018}
}

@article{open,
  title={Towards open vocabulary learning: A survey},
  author={Wu, Jianzong and Li, Xiangtai and Xu, Shilin and Yuan, Haobo and Ding, Henghui and Yang, Yibo and Li, Xia and Zhang, Jiangning and Tong, Yunhai and Jiang, Xudong and others},
  journal={IEEE Transactions on Pattern Analysis and Machine Intelligence},
  volume={46},
  number={7},
  pages={5092--5113},
  year={2024},
  publisher={IEEE}
}
}

\end{document}